\documentclass{article}

\usepackage{arxiv}

\usepackage[utf8]{inputenc} 
\usepackage[T1]{fontenc}    
\usepackage{hyperref}       
\usepackage{url}            
\usepackage{booktabs}       
\usepackage{amsfonts}       
\usepackage{nicefrac}       
\usepackage{microtype}      
\usepackage{lipsum}
\usepackage{graphicx}
\usepackage{array}
\usepackage{multirow}
\usepackage{ragged2e}
\usepackage{booktabs}
\usepackage[table]{xcolor}
\graphicspath{ {./images/} }
\usepackage{amsmath}

\title{An Efficient and Mixed Heterogeneous Model for Image Restoration}

\author{
\textbf{Yubin Gu}$^{1,}$\thanks{Equal contribution.}, \textbf{Yuan Meng}$^{1,}$\footnotemark[1], \textbf{Kaihang Zheng}$^{1,}$\footnotemark[1], \textbf{Xiaoshuai Sun}$^{1,}$\thanks{Corresponding author.}  \\
\textbf{Jiayi Ji}$^{1,2}$, \textbf{Weijian Ruan}$^{3}$, \textbf{Liujuan Cao}$^{1}$, \textbf{Rongrong Ji}$^{1}$ \\
\\
$^1$MAC Lab, Xiamen University, China \\
$^2$National University of Singapore, Singapore \\
$^3$Smart City Research Institute, China Electronics Technology Group Corporation, China
}

\begin{document}
\maketitle
\begin{abstract}
  Image restoration~(IR), as a fundamental multimedia data processing task, has a significant impact on downstream visual applications. In recent years, researchers have focused on developing general-purpose IR models capable of handling diverse degradation types, thereby reducing the cost and complexity of model development. Current mainstream approaches are based on three architectural paradigms: CNNs, Transformers, and Mambas. CNNs excel in efficient inference, whereas Transformers and Mamba excel at capturing long-range dependencies and modeling global contexts. While each architecture has demonstrated success in specialized, single-task settings, limited efforts have been made to effectively integrate heterogeneous architectures to jointly address diverse IR challenges. To bridge this gap, we propose RestorMixer, an efficient and general-purpose IR model based on mixed-architecture fusion. RestorMixer adopts a three-stage encoder-decoder structure, where each stage is tailored to the resolution and feature characteristics of the input. In the initial high-resolution stage, CNN-based blocks are employed to rapidly extract shallow local features. In the subsequent stages, we integrate a refined multi-directional scanning Mamba module with a multi-scale window-based self-attention mechanism. This hierarchical and adaptive design enables the model to leverage the strengths of CNNs in local feature extraction, Mamba in global context modeling, and attention mechanisms in dynamic feature refinement. Extensive experimental results demonstrate that RestorMixer achieves leading performance across multiple IR tasks while maintaining high inference efficiency. The official code can be accessed at \url{https://github.com/ClimBin/RestorMixer}.
\end{abstract}

\section{Introduction}

Image restoration (IR) is a fundamental task in multimedia visual data processing, aimed at recovering high-quality, clear images from degraded or damaged images. As a critical pre-processing step, IR significantly affects the performance of downstream visual applications, such as object detection~\cite{liu2024unsupervised_dete}, image segmentation~\cite{wang2024cascaded_seg}, and video analysis~\cite{tan2024blind_video1}. 
Therefore, IR has become an integral component of modern visual system data processing pipelines.

\begin{figure}[t]
  \centering
  \includegraphics[width=0.56\linewidth]{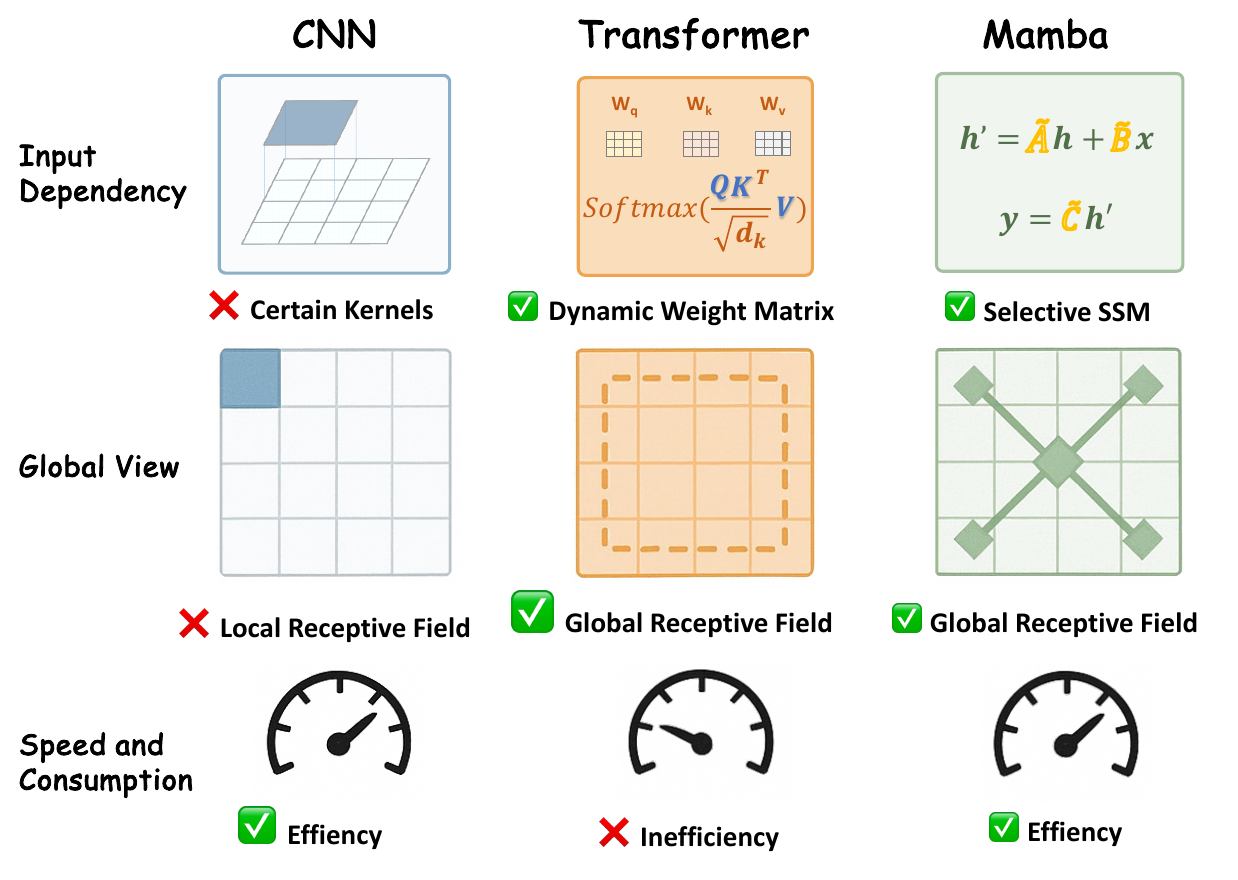}
  \caption{Comparison of three mainstream architectures in terms of input dependency, global view, and inference speed. }
  \label{fig:intro}
\end{figure}
Traditionally, IR models have been designed for specific tasks, such as deraining~\cite{zou2024freqmamba,IDT}, desnowing~\cite{zhang2023hcsd_snow,chen2022snowformer}, and super-resolution (SR)~\cite{zhan2024anysr,dong2015image_srcnn}. While these task-specific models typically achieve excellent performance, they are costly in terms of development and deployment. Each task requires specialized architecture design and corresponding model settings, which lowers efficiency in both research and practical applications. With the rapid development of deep learning technologies, there has been increasing interest in developing general-purpose IR models capable of handling multiple types of degradation within the same types of frameworks. These models have shown a tendency to either approach or even surpass the performance of task-specific expert models. Such models not only reduce design and deployment costs but also promote scalability and adaptability across diverse IR tasks. Thus, general-purpose IR models represent a meaningful and promising research direction, as they can significantly lower development costs while achieving performance comparable to expert models.

Recent works have proposed general-purpose IR models, primarily focusing on three major architectures: Convolutional Neural Networks (CNN), Transformers, and Mamba-based architecture. Fig.~\ref{fig:intro} shows the advantages of different design mechanisms.
For instance, the CNN-based IRNeXt~\cite{cui2023irnext} leverages the efficiency and locality of convolutions to effectively extract shallow features. Restormer~\cite{zamir2022restormer} and SwinIR~\cite{liang2021swinir_swinir} are Transformer-based models that utilize various self-attention mechanisms to model global feature long-range dependencies and window-granularity feature relationships, enhancing the model's dynamic perception of features. These models outperform earlier CNN-based models, though at the cost of slower inference speed. Additionally, the recently proposed Mamba~\cite{gu2023mamba} model incorporates unique hardware-aware and parallel scanning mechanisms that accelerate global modeling, offering a new foundational structure for IR models. For example, VmambaIR~\cite{shi2024vmambair} and MambaIR~\cite{guo2024mambair} introduce innovative scanning mechanisms to establish global dependencies of visual features in multiple directions, addressing limitations of the vanilla Mamba in handling visual data with non-causal relationships. However, they overlook issues of feature redundancy and computational deficiency caused by multi-directional scanning and memory loss due to long sequences. These methods offer diverse perspectives on general-purpose IR design, focusing on single-structure-based solutions, yet they overlook the potential benefits of integrating heterogeneous designs to address the diversified challenges of IR tasks. Consequently, existing general-purpose IR models often exhibit suboptimal performance due to the inability to fully exploit the complementary advantages of different architectural designs.
\begin{figure}[t]
    \centering
    \includegraphics[width=.5\linewidth]{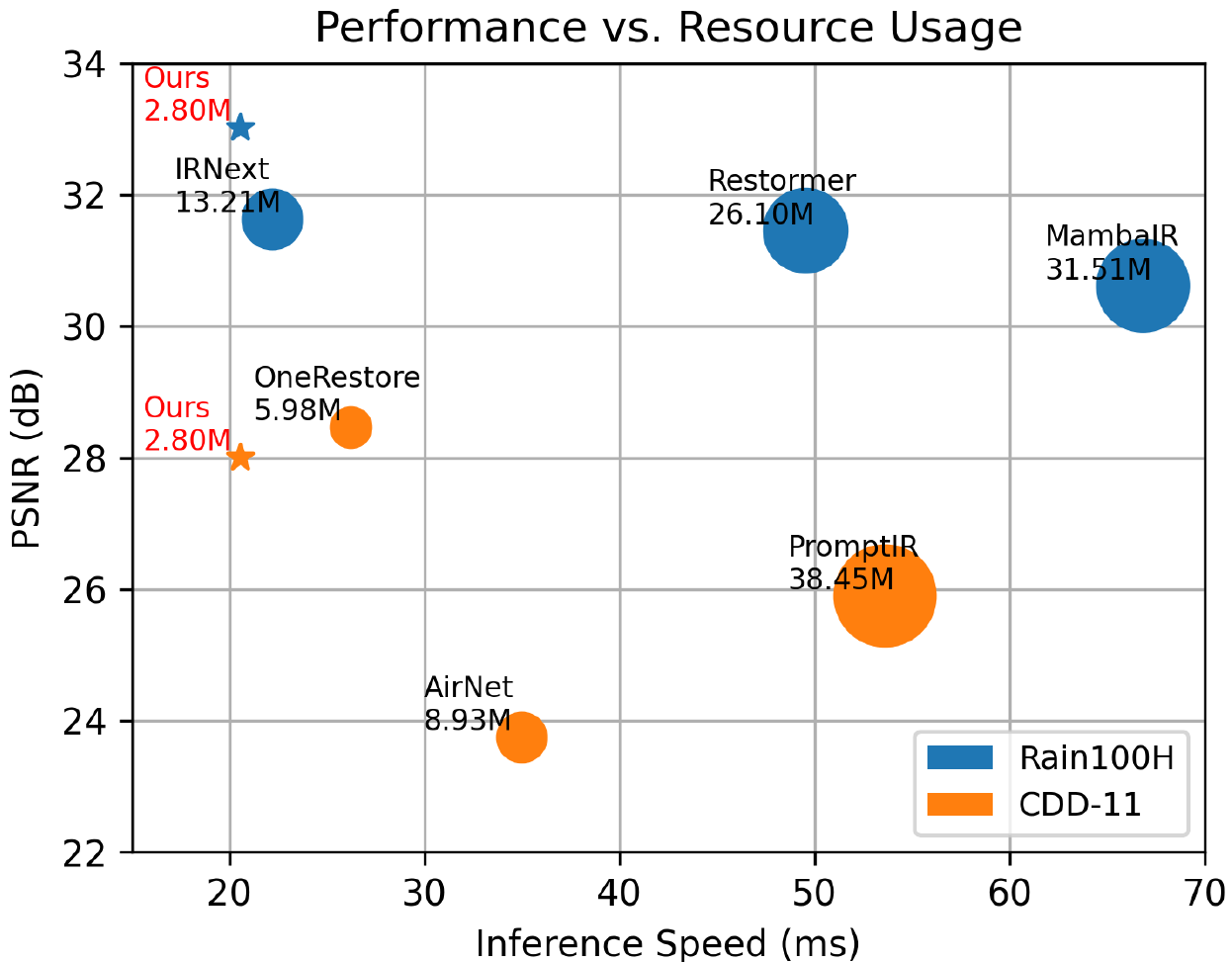}
    \caption{Comparison of RestorMixer with various representative methods in terms of performance, inference speed, and number of parameters. Testing is at the same input size.}
    \label{fig:infers}
\end{figure}


To address the research gap in heterogeneous architecture design for general-purpose IR models, we propose RestorMixer - a novel hierarchical framework integrating complementary architectural paradigms. The model employs a three-stage encoder-decoder architecture with progressive downsampling/upsampling operations for multi-scale feature learning. Each stage is specifically optimized according to inherent scale characteristics: In the high-resolution initial stage, lightweight convolutional blocks efficiently extract shallow features through local receptive fields, prioritizing fine-grained detail preservation with minimal inference overhead. Subsequent stages implement a mixed architecture M-T blocks combining: 1) Enhanced Memory Visual Mamba (EMVM) blocks for global dependency modeling, and 2) Multi-scale Window-based Transformer blocks that dynamically refine multi-ranged local representations through attention-guided feature recalibration. This strategic integration synergizes CNN's local inductive bias, Mamba's linear-complexity global modeling, and Transformer's adaptive feature optimization. Experimental validation demonstrates that RestorMixer achieves leading performance on both single/mixed degradation benchmarks while maintaining superior inference efficiency, as shown in Fig.~\ref{fig:infers}. In general, our contributions can be summarized as follows:
\begin{itemize}
    \item We analyze the limitations of existing general-purpose IR models that focus on single-architecture design paradigms and emphasize the necessity of adopting a mixed structure fusion approach to address the diverse requirements of IR.
    \item We propose RestorMixer, a multi-stage encoder-decoder IR model that effectively integrates CNN, Mamba, and Transformer. This design not only leverages the strengths of each design but also optimizes their combination and collaboration based on the characteristics of features. 
    \item Through extensive experiments on various benchmarks, including single-degradation and mixed-degradation datasets, we demonstrate that RestorMixer achieves leading performance while maintaining efficient inference.
\end{itemize}

\section{Related Work}
\subsection{Image Restoration}
Image Restoration (IR) seeks to recover high-quality images from degraded inputs. It is a central challenge in multimedia data processing due to its ill-posed nature and diverse subtasks. Over the past decade, research has shifted from traditional handcrafted feature extraction methods to data-driven deep neural network-based learning models, which have become mainstream due to their superior performance.
Researchers have explored multiple subtasks, such as rain streak removal~\cite{zou2024freqmamba}, super-resolution~\cite{zhan2024anysr}, and mixed degradation IR~\cite{guo2024onerestore} et al. For example, Wang et al.~\cite{wang2024progressive_plni} proposed a model for rain removal using local and non-local interactions and multi-scale deep learning. In super-resolution, SRCNN~\cite{dong2015image_srcnn} pioneered the use of deep learning with a shallow convolutional neural network. Besides, Zhan et al.'s AnySR~\cite{zhan2024anysr} further reconstructs arbitrary-scale super-resolution methods for flexible resource allocation.
However, specialized models for specific degradations have limited generalizability. Recent research has focused on developing general IR models~\cite{cui2023irnext,zamir2022restormer,liang2021swinir_swinir,guo2024mambair,wang2022uformer} to handle multiple tasks with a single architecture. Existing general-purpose IR models are based on single architectural frameworks like CNNs, Transformers, and state space models (e.g., Mamba). CNN-based methods like MPRNet~\cite{MPRNet} and IRNeXt~\cite{cui2023irnext} offer robust feature extraction for universal restoration. Transformer-based models like Restormer~\cite{zamir2022restormer} and SwinIR~\cite{liang2021swinir_swinir} capture long-range dependencies but face computational complexity issues. Recently, state space models such as Mamba~\cite{gu2023mamba} have gained attention for their computation ability and long-sequence modeling capabilities. Although CNNs have distinct advantages in inference speed, Transformers in global modeling, and Mamba in efficient long-range modeling, few studies have explored the integration of these triple heterogeneous foundational architectures for general-purpose IR. 

\begin{figure*}[t]
  \centering
  \includegraphics[width=1\linewidth]{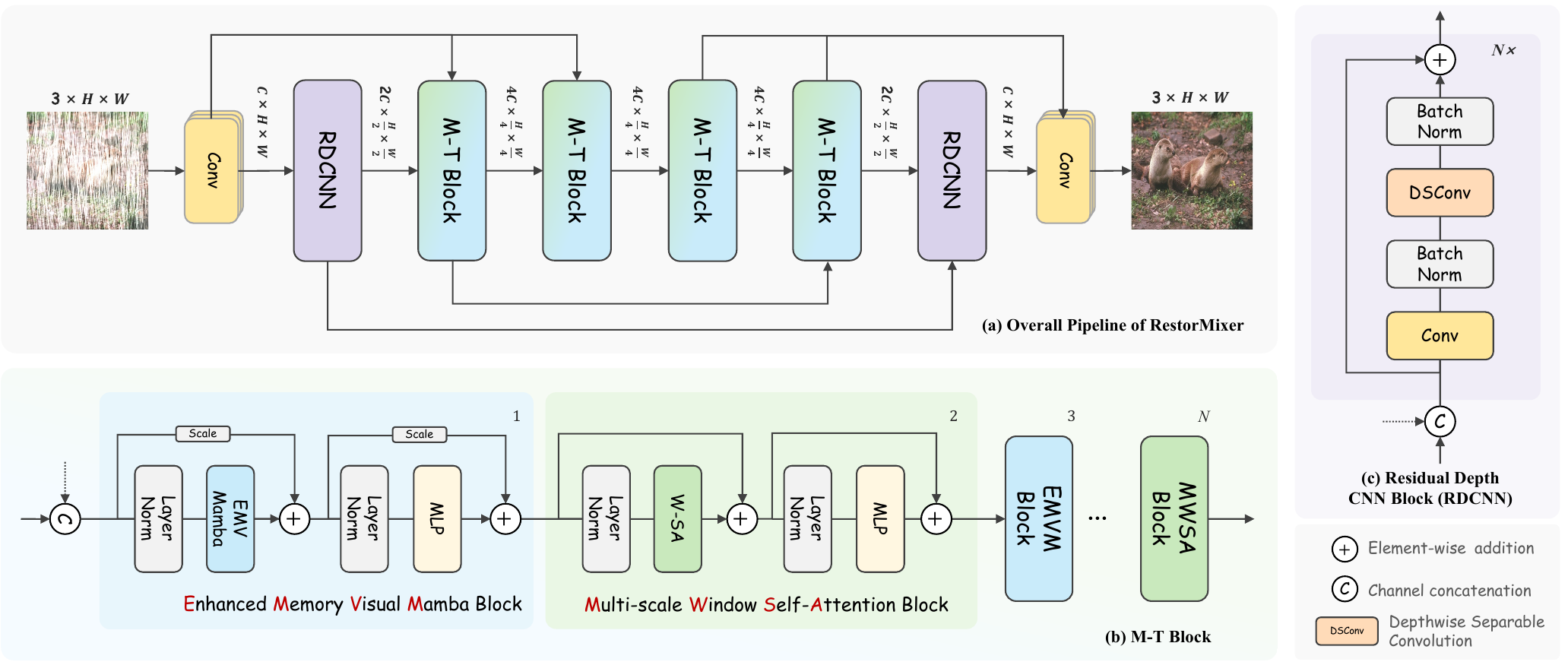}
  \caption{Framework of \textbf{RestorMixer}. (a) Overall pipeline. (b) Structure of the M-T Blocks, composed of alternating Enhanced Memory Visual Mamba Blocks (EMVM) and Multi-scale Window Self-Attention (MWSA) Blocks. (c) Residual Depth CNN Block (RDCNN), built with a stack of basic residual convolutional units.}
  \label{fig:pipeline}
\end{figure*}
\subsection{State Space Models}

State space models (SSMs), originally inspired by classical control theory, have emerged as a novel backbone in deep learning, particularly excelling in modeling long-sequence data. By incorporating linear extrapolation properties, these models have achieved breakthrough progress in long-range dependency modeling, attracting significant research interest.
Notably, Mamba~\cite{gu2023mamba}, a new generation of input-dependent SSMs, has demonstrated Transformer-like performance in natural language processing (NLP), with computational complexity linear in input length, thanks to its selective mechanism and hardware-efficient ability. The success of Mamba has spurred its application in vision tasks, including image classification~\cite{hatamizadeh2024hybrid_mambavision} and biomedical image segmentation~\cite{ma2024u_mamba}. 
Recent innovations~\cite{shi2024multis_vm,hatamizadeh2024hybrid_mambavision,han2024demystify_demamb,shi2024vmambair} based on Vision Mamba~\cite{liu2024vmamba} have shown promise. 
In the IR domain, models such as MambaIR~\cite{guo2024mambair} and VMambaIR~\cite{shi2024vmambair} have demonstrated significant potential. 
For example, MambaIR~\cite{guo2024mambair} combined visual SSMs with improved MLP to address local pixel forgetting and channel redundancy in traditional Mamba.
However, Mamba's unidirectional scanning mechanism limits its performance in image data processing and in-context learning compared to Transformers. Although multi-directional scanning mechanisms have mitigated some of these limitations for non-causal image data, they introduce computational redundancy and insufficient local feature association. To address these challenges, this work introduces an optimized scanning mechanism in the encoder-decoder's latter two low-resolution feature processing stages, combining Mamba with a simple multi-scale window-based Transformer structure. This approach effectively resolves issues of computational redundancy and insufficient local feature association.

\section{The Proposed Model: RestorMixer}

\subsection{Overview of Architecture}
We introduce RestorMixer, a novel IR model that integrates heterogeneous structures. Fig.~\ref{fig:pipeline}~(a) shows that RestorMixer differs from prior single-structure models (\textit{e.g.}, CNN, Transformer, Vision Mamba) by combining three foundational designs tailored to the feature properties processed at different encoder/decoder stages, ensuring synergistic interaction.

The model comprises two main blocks: the Residual Depth CNN Block (RDCNN) and the Mamba-Transformer Block (M-T), distinguished by colored blocks in Fig.~\ref{fig:pipeline}~(a). Given a degraded input image \(\mathcal{I}_{low} \in \mathbb{R}^{H\times W \times C}\), the model first adopts a CNN stem to expand input channels, yielding \(\mathcal{I}_{low}' \in \mathbb{R}^{H\times W \times C}\) (with \(C = 32\) channels). The encoding process involves three stages, each handling features at different scales. Specifically, \(\mathcal{I}'\) passes through the Residual Depth CNN Block~(RDCNN), as shown in Fig.~\ref{fig:pipeline}~(c) to extract shallow features \(\mathcal{I}_{e}^1 \in \mathbb{R}^{H\times W\times C}\), followed by a convolutional downsampling layer that halves the spatial scale and doubles the channels to produce \(\mathcal{I}_{e}^2 \in \mathbb{R}^{\frac{H}{2}\times \frac{W}{2} \times 2C}\). \(\mathcal{I}_{e}^2\) is then concatenated with the corresponding downsampled input features \(\mathcal{I}_{/2}' \in \mathbb{R}^{\frac{H}{2} \times \frac{W}{2}\times C}\) and fed into the M-T Block (as shown in Fig.~\ref{fig:pipeline}~(b)). This process continues to generate three multi-scale feature maps in the encoder. In the decoder, M-T Blocks process lower-resolution feature maps, while RDCNN block handles the highest-resolution features. Additionally, we employ a deep supervision strategy during optimization, enabling each decoder stage to produce predictions at corresponding scales for loss calculation against reference images. Next, we detail the three foundational blocks comprising RestorMixer.

\begin{figure}[t]
  \centering
  \includegraphics[width=.65\linewidth]{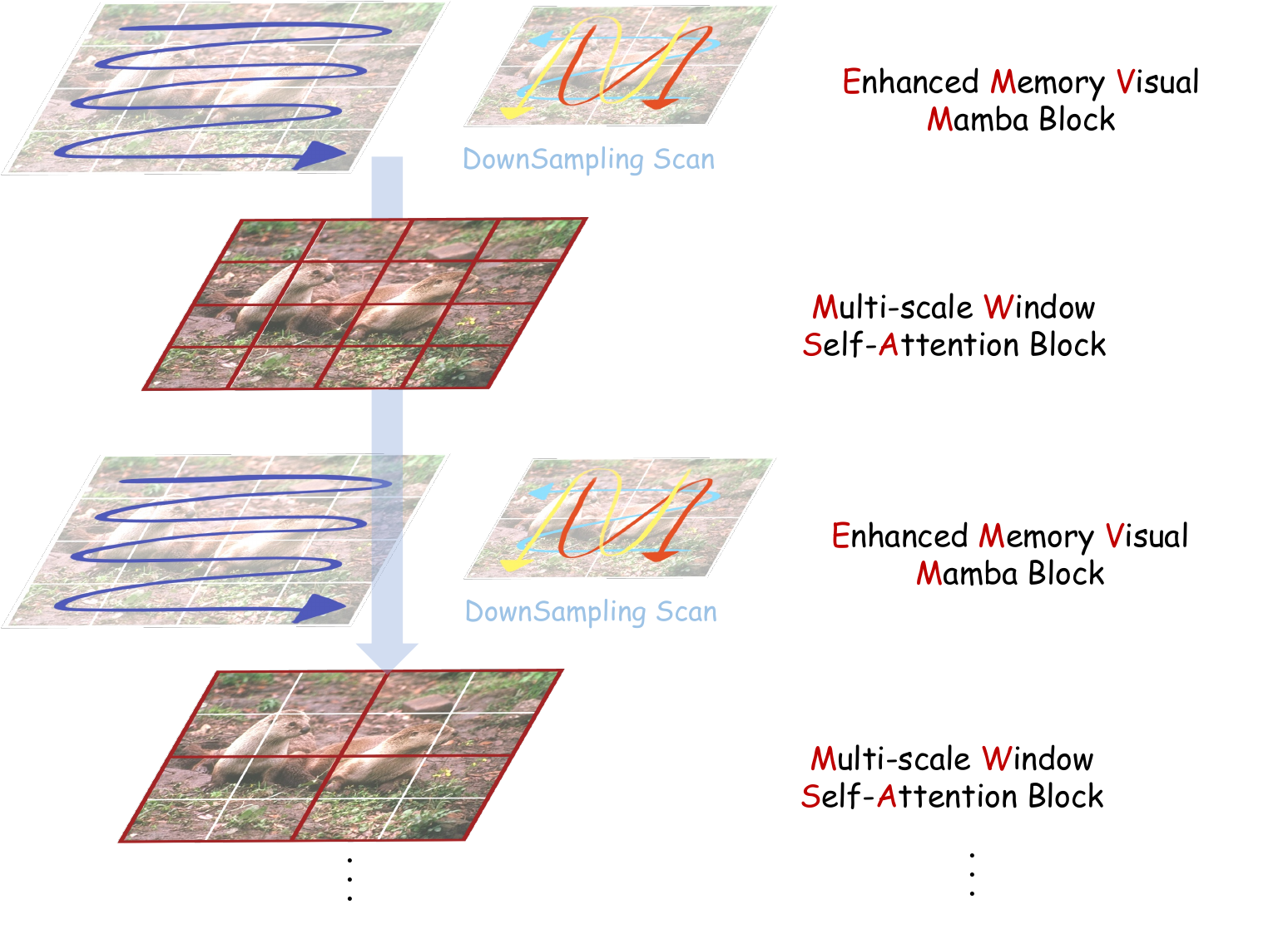}
  \caption{Illustration of the M-T Block. It is constructed by alternately stacking EMVM blocks with four-directional scanning and MWSA blocks to jointly capture long-range dependencies and local multi-scale features.}
  \label{fig:MT}
\end{figure}

\subsection{The Heterogeneous Designed Modules}
RestorMixer combines three main heterogeneous structures with a focus on enhancing module collaboration. For large feature maps in high-resolution spaces, a rational RDCNN block is used due to its efficiency in extracting shallow features and parallel processing capability. For two lower-resolution stages, modules based on Mamba and Transformer architectures are alternated to model long-range dependencies and refine local representations.
\subsubsection{\textbf{Residual Depth CNN Block}}
We designed the Residual Depth CNN Block (RDCNN) to process first-stage input features, which integrates standard convolution and depthwise separable convolution, as shown in Fig.~\ref{fig:pipeline}~(c). For an input feature map \( \mathbf{X} \in \mathbb{R}^{H \times W \times C} \), the spatial feature extraction is formulated as:  
\begin{equation}
    \mathbf{Y}_1 = \text{GELU}\left(\text{BN}\left(\text{Conv}_{3\times3}(\mathbf{X})\right)\right),    
\end{equation}
where BN($\cdot$) and GELU($\cdot$) are the Batch Normalization and the activation function, respectively. Standard \( 3\times3 \) convolution captures fine-grained local textures and high-frequency patterns critical for IR, such as edge reconstruction and noise suppression. Subsequent processing employs a depthwise separable convolution to decouple spatial filtering and channel interaction:  
\begin{equation}
    \mathbf{Y}_{dw} = \text{Depth-Conv}_{3\times3}(\mathbf{Y}_1), \quad \mathbf{Y}_{pw} = \text{Point-Conv}_{1\times1}(\mathbf{Y}_{dw}),    
\end{equation}
which reduce the computational complexity from \( \mathcal{O}(C_{\text{in}}C_{\text{out}}K^2) \) to \( \mathcal{O}(C_{\text{out}}(K^2 + C_{\text{in}})) \). This design leverages the complementary strengths of both operations: the standard convolution preserves spatially correlated details essential for pixel-wise restoration, while the depthwise separable convolution minimizes redundant parameters, enhancing efficiency for large-scale feature maps of the first stage. The residual connection:  
\begin{equation}
    \mathbf{Y}_{\text{out}} = \text{GELU}\left(\text{BN}(\mathbf{Y}_{pw}) + \mathbf{X}\right),    
\end{equation}
ensures stable gradient flow and feature reuse, mitigating information loss in the IR model. This module is located at the beginning of the encoding stage and the end of the decoding stage to efficiently process the feature map of the largest scale, effectively extract and express shallow features, and both provide favorable conditions for the subsequent encoding stage and the final prediction output.

\subsubsection{\textbf{Enhanced Memory Visual Mamba}}
Initially, we provide a brief introduction to the State Space Models (SSM). SSMs map an input sequence \(x(t) \in \mathbb{R}^L\) to a latent representation \(h(t) \in \mathbb{R}^N\) and predict an output sequence \(y(t) \in \mathbb{R}^L\). Mathematically, they are described by:
\begin{equation}
    h'(t) = \mathcal{A}h(t) + \mathcal{B}x(t), \quad y(t) = \mathcal{C}h(t),    
\end{equation}
where \(A \in \mathbb{R}^{N \times N}\), \(B \in \mathbb{R}^{N \times 1}\), and \(C \in \mathbb{R}^{1 \times N}\) are learnable parameters. To adapt SSMs for deep learning frameworks, discretization is applied using a timescale parameter \(\Delta \in \mathbb{R}\) and the zero-order hold (ZOH) rule. The discretized equations become:
\begin{equation}
    h(t) = \mathcal{\bar A}h(t-1) + \mathcal{\bar B}x(t), \quad y(t) = \mathcal{C}h(t),    
\end{equation}
where \(\mathcal{\bar A} = e^{\Delta \mathcal{A}}\) and \(\mathcal{\bar B} = (\Delta \mathcal{A})^{-1}(e^{\Delta \mathcal{A}} - I)\Delta \mathcal{B} \approx \Delta \mathcal{B}\), with \(I\) as the identity matrix. This process can be implemented as a global convolution:
\begin{equation}
    y = x \odot \mathcal{K}, \quad \mathcal{K} = [\mathcal{CB, CAB}, \dots, \mathcal{CA}^{L-1}\mathcal{B}],    
\end{equation}
where \(\mathcal{K} \in \mathbb{R}^L\) is the convolution kernel.
Furthermore, the Selective State Space mechanism equipped with the popular Mamba model enhances SSMs by making parameters \(\mathcal{B}\), \(\mathcal{C}\), and \(\Delta\) input-dependent. Such a design enhances the model's input perception ability.

Recent IR methods have incorporated multi-directional scanning mechanisms to leverage Mamba's efficient sequence modeling capabilities and address the lack of multi-perspective modeling at the single-pixel level in vanilla Mamba. While this approach mitigates issues with uni-directional scanning in visual data processing, it introduces redundant computations and memory decay in single scans. To address these challenges, we propose the Enhanced Memory Visual Mamba (EMVM) module, as illustrated in Fig.~\ref{fig:pipeline}~(b, blue block) and Fig.~\ref{fig:MT}. EMVM adopts a multi-directional scanning strategy, processing visual sequences through horizontal forward (hf), horizontal backward (hb), vertical forward (vf), and vertical backward (vb) scans. Unlike previous methods~\cite{guo2024mambair,shi2024vmambair}, EMVM optimizes for single-direction forgetting and reduces redundant computations across multiple directions, thereby enhancing model performance.

Specifically, given an \( \mathbf{X} \in \mathbb{R}^{H\times W\times C}\), EMVM first performs Layer Normalization on the \(\mathbf{X}\) to obtain \(\mathbf{X'}\):
\begin{equation}
    \mathbf{X'} = \text{LayerNorm}(\mathbf{X}),
\end{equation}
Then, \(\mathbf{X'}\) is reorganized in the horizontal forward direction to obtain \( \mathbf{X_{hf}} \in \mathbb{R}^{HW\times C}\), followed by the selective state-space modeling to capture the transformation patterns for IR: 
\begin{equation}
    \mathbf{Y_{hf}} = {\text{S}^{3}\text{M}}(\mathbf{X_{hf}}) = \text{SelectiveSSM}\left(\text{Linear}(\mathbf{X_{hf}})\right), 
\end{equation}
where the state-space dynamics adapt to input content through discretized parameters \( \bar{\mathcal{A}}, \bar{\mathcal{B}} \):  
\begin{equation}
    h_t = \bar{\mathcal{A}} \odot h_{t-1} + \bar{\mathcal{B}} \odot x'_t.      
\end{equation}
Besides, to reduce computational redundancy in multi-directional scanning and enhance scanning memory, we first downsample the input feature map \(\mathbf{X'}\) by a factor of 2. Then, we re-sequence the data according to specific directions to obtain \(\mathbf{X}_{\text{hb}} \in \mathbb{R}^{\frac{H}{2}\times \frac{W}{2}\times C}\), \(\mathbf{X}_{\text{vf}}\in \mathbb{R}^{\frac{H}{2}\times \frac{W}{2}\times C}\), and \(\mathbf{X}_{\text{vb}}\in \mathbb{R}^{\frac{H}{2}\times \frac{W}{2}\times C}\). Each directional sequence is modeled similarly to \(\mathbf{X}_{\text{hf}}'\):
\begin{equation}
    \{\mathbf{X_{hb},X_{vf},X_{vb}}\} = \text{Downsampling}(\mathbf{X'}),
\end{equation}
\begin{equation}
    \{\mathbf{Y_{hb}',Y_{vf}',Y_{vb}'}\} \to  \{{\text{S}^{3}\text{M}}(\mathbf{X_{hb}}),{\text{S}^{3}\text{M}}(\mathbf{X_{vf}}),{\text{S}^{3}\text{M}}(\mathbf{X_{vb}})\},
\end{equation}
then followed by upsampling by a factor of 2 to restore the original feature scale, yielding \(\mathbf{Y}_{\text{hb}}\), \(\mathbf{Y}_{\text{vf}}\), and \(\mathbf{Y}_{\text{vb}}\). Finally, the outputs from all four directions are summed to produce the final output \(\mathbf{Y}\):
\begin{equation}
    \mathbf{Y} = \mathbf{Y_{hf}} + \mathbf{Y_{hb}} + \mathbf{Y_{vf}} + \mathbf{Y_{vb}}.
\end{equation}
After getting the Mamba's output, a learnable channel-wise scaling factor \( \gamma_1 \) preserves critical spatial details during residual fusion:  
\begin{equation}
    \mathbf{Z} = \gamma_1 \odot \mathbf{X} + \mathbf{Y}.      
\end{equation}
Local texture refinement is then achieved through MLP-enhanced feature modulation:  
\begin{equation}
    \hat{\mathbf{Z}} = \gamma_2 \odot \mathbf{Z} + \text{MLP}(\text{LayerNorm}(\mathbf{Z})),      
\end{equation}
where \( \gamma_2 \) further recalibrates channel importance. Finally, $\mathbf{Z}$ passes through an MLP layer equipped with weighted residual connections to produce the output $\hat{\mathbf{Z}\in\mathbb{R}^{H\times W\times C}}$ of EMVM, enhancing feature extraction and representation capabilities.

\subsubsection{\textbf{Multi-scale Window Self-attention}}
Furthermore, we designed a multi-scale window self-attention module (MWSA), as shown in Fig.~\ref{fig:pipeline}~(b) and Fig.~\ref{fig:MT}, which dynamically refines local feature representations across hierarchical receptive fields to complement the global dependency modeling of EMVM. Designed to address the dual demands of high-frequency detail preservation and computational efficiency, MWSA processes an input feature map \(\mathbf{X} \in \mathbb{R}^{H \times W \times C}\) through window-based self-attention with progressively expanding window sizes. For a target window size \(w_s\), the input is partitioned into \(w_s \times w_s\) windows \(\{\mathbf{W}_{i,j}\}\), each transformed via layer normalization and self-attention:  
\begin{equation}
    \mathbf{W}'_{i,j} = \text{LayerNorm}(\mathbf{W}_{i,j}),    
\end{equation}
\begin{equation}
    \text{Attn}(\mathbf{Q},\mathbf{K},\mathbf{V}) = \text{Softmax}\left(\frac{\mathbf{Q}\mathbf{K}^T}{\sqrt{d_k}}\right)\mathbf{V},    
\end{equation}
where \(\mathbf{Q} = \mathbf{W}'_{i,j}\mathbf{W}_Q\), \(\mathbf{K} = \mathbf{W}'_{i,j}\mathbf{W}_K\), \(\mathbf{V} = \mathbf{W}'_{i,j}\mathbf{W}_V\) are linear projections. The attended features are reconstructed into the spatial domain and fused with the original input through residual connections:  
\begin{equation}
    \mathbf{X}_{\text{attn}} = \text{WindowMerge}(\{\mathbf{W}'_{i,j}\}), \quad \mathbf{X}' = \mathbf{X} + \mathbf{X}_{\text{attn}},    
\end{equation}
followed by an MLP for nonlinear refinement:  
\begin{equation}
    \mathbf{X}_{\text{out}} = \mathbf{X}' + \text{MLP}(\text{LayerNorm}(\mathbf{X}')).    
\end{equation}
By incrementally expanding window sizes from \(8\times8\) to larger scales (\(+8\) per module), MWSA hierarchically aggregates local-to-semiglobal contexts—small windows enhance details, while larger windows capture structural coherence. 
The design reduces global attention complexity, enabling efficient interaction between EMVM and MWSA. Their alternating use ensures integration of global recovery and multi-scale detail reconstruction, crucial for precise and efficient IR tasks.

\subsection{Optimization Objective}

For the super-resolution, we adopted single-scale L1 loss function to ensure fair comparison with existing studies~\cite{gu2023mamba}:
\begin{equation}
\mathcal{L}_{\text{SR}} = \| I_{\text{HR}} - \hat I_{\text{HR}} \|_1,
\end{equation}
where \(I_{\text{HR}}\) and \(\hat I_{\text{HR}}\) denote high-resolution ground truth and reconstructed high-resolution image respectively. This single-scale formulation aligns with conventional evaluation protocols.
For other tasks, we implemented deep supervision with multi-scale outputs. The composite loss integrates spatial and frequency domain constraints across scales:
\begin{equation}
\mathcal{L}_{\text{total}} = \sum_{i=1}^{3} \frac{1}{N_i} \left( \| P_i - I_i \|_1 + \lambda \| \mathcal{F}(P_i) - \mathcal{F}(I_i) \|_1 \right),
\end{equation}
where \(P_i\) and \(I_i\) represent the predicted and target images at scale \(i\), \(N_i\) is pixel count, and \(\mathcal{F}(\cdot)\) denotes Fourier transform. The \(\lambda=0.1\) balance factor follows established practices~\cite{cui2023irnext}. This multi-scale dual-domain design enhances feature learning for improved restoration quality.

\section{Experiments and Analysis}
\subsection{Datasets and Experimental Settings}

To comprehensively assess RestorMixer across diverse degradation scenarios, we conduct evaluations on both single and composite image restoration tasks. The single degradation tasks include deraining, desnowing and super-resolution. For deraining, we use Rain100L~\cite{100HL} (light rain), Rain100H~\cite{100HL} (heavy rain), Test1200~\cite{TEST1200DID} (dense mild rain), and Test2800~\cite{TEST2800} (multi-directional rain). For desnowing, we evaluate on the CSD~\cite{csd} dataset. Super-resolution models are trained on DIV2K~\cite{DIV2K} with upscaling factors of $\times$2, $\times$3, and $\times$4, and tested on Set5~\cite{set5}, Set14~\cite{set14}, BSDS100~\cite{bsds100}, Urban100~\cite{urban100}, and Manga109~\cite{manga109}. Composite degradation is trained and evaluated on the CDD-11~\cite{guo2024onerestore} benchmark, which includes 4 single degradations (Low, Haze, Rain, Snow), 4 dual-combinations (Low + Haze, Low + Rain, Low + Snow, Haze + Rain), and 3 triple-combinations (Low+Haze+Rain, Haze+Snow, Low+Haze+Snow).

RestorMixer adopts a unified 3-stage encoder-decoder architecture with 4 basic blocks per stage for most tasks. For super-resolution, Adjustments of the model are in the \textit{Supplementary Material}. 
All models are trained using the Adam~\cite{kingma2014adam} optimizer, and data augmentation includes random cropping and horizontal flipping. The learning rate follows a cosine annealing schedule with task-specific initialization.
Following common practice, PSNR and SSIM are used for evaluation. Following the previous works, deraining results are reported in YCbCr and all others in RGB. All experiments are run on NVIDIA RTX 4090 GPUs. 

\subsection{Rainy Streak Removal of Single Image}



\begin{figure*}[t]
  \centering
  \includegraphics[width=1\linewidth]{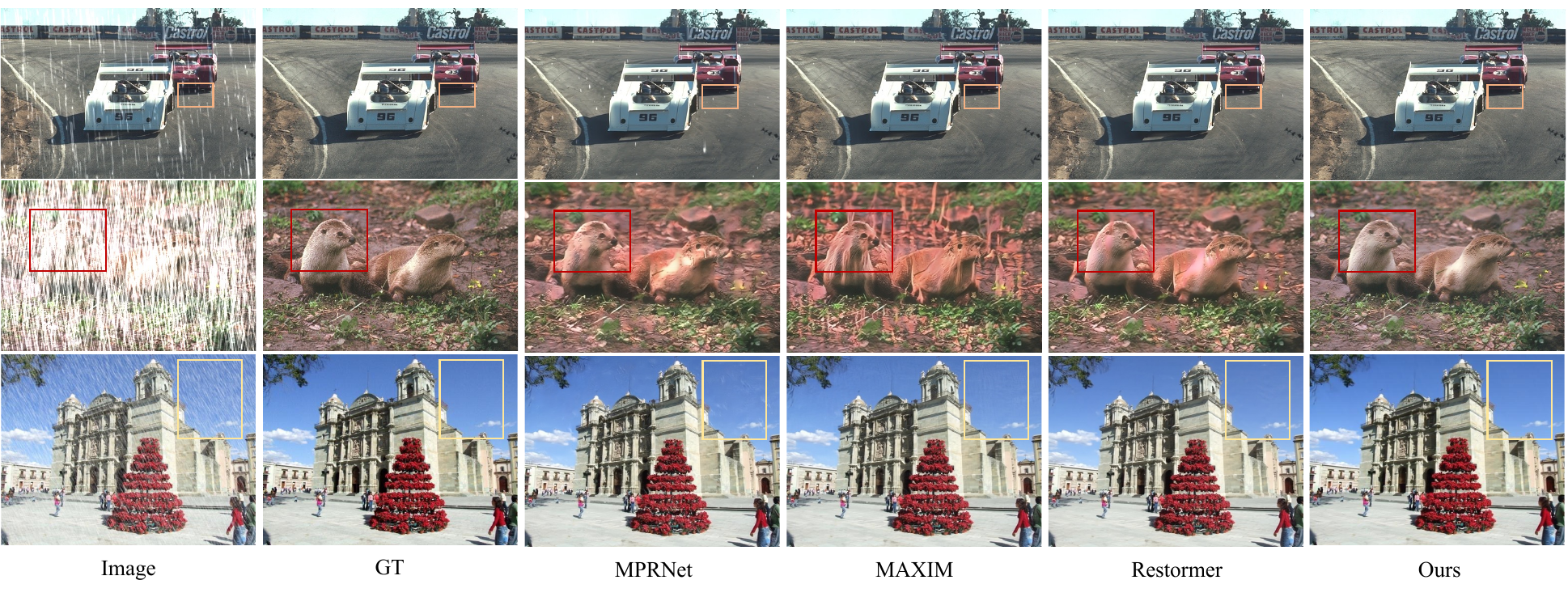}
  \caption{Visual comparison with SOTA methods under various rain intensities.}
  \label{fig:rain}
\end{figure*}

RestorMixer achieves state-of-the-art performance on four deraining benchmarks: Rain100H~\cite{100HL}, Rain100L~\cite{100HL}, Test2800~\cite{TEST2800}, and Test1200~\cite{TEST1200DID}. As shown in Tab.~\ref{table:rain}, it obtains an average PSNR of 35.47 dB and SSIM of 0.949, clearly outperforming existing methods. The visual results in Fig.~\ref{fig:rain} confirm its advantage in restoring details and removing rain streaks.

On the challenging Test1200 dataset, RestorMixer achieves a PSNR of 34.93\,dB, surpassing the second-best method by 1.57\,dB. In heavy rain scenarios such as Rain100H, the improvement reaches 1.29 dB, highlighting its strong capacity to model global and local features, as well as its generalization of diverse rain patterns.  Despite its strong performance, RestorMixer maintains a lightweight design with only 2.80M parameters and 25.16G FLOPs, the lowest among all compared methods, demonstrating an excellent trade-off between accuracy and efficiency, and strong potential for deployment in real-world, resource-constrained environments.

\begin{table*}[t]
\centering
\renewcommand{\arraystretch}{1.2}
\setlength{\tabcolsep}{3.3pt}

\caption{Quantitative comparison of various methods on rain streak removal on four public rain datasets.}
\label{table:rain}
\begin{tabular}{lcccccccccccc} 
\toprule
\multirow{2}{*}{\textbf{Methods}} & \multicolumn{2}{c}{\textbf{Rain100H}} & \multicolumn{2}{c}{\textbf{Rain100L}} & \multicolumn{2}{c}{\textbf{Test2800}} & \multicolumn{2}{c}{\textbf{Test1200}} & \multicolumn{2}{c}{\textbf{Average}} & \multirow{2}{*}{\begin{tabular}[c]{@{}c@{}}\textbf{Params}\\\textbf{(M)}\end{tabular}} & \multirow{2}{*}{\begin{tabular}[c]{@{}c@{}}\textbf{Flops}\\\textbf{(G)}\end{tabular}}  \\
& \textbf{PSNR}  & \textbf{SSIM}  & \textbf{PSNR}  & \textbf{SSIM}  & \textbf{PSNR}  & \textbf{SSIM}  & \textbf{PSNR}  & \textbf{SSIM}  & \textbf{PSNR}  & \textbf{SSIM}  &        &         \\

\midrule
MSPFN~\cite{MSPFN}           & 28.66 & 0.860 & 32.40 & 0.933 & 32.82 & 0.930 & 32.39 & 0.916 & 31.57 & 0.910 & 13.22 & 604.70 \\
MPRNet~\cite{MPRNet}          & 30.41 & 0.890 & 36.40 & 0.965 & 33.64 & 0.938 & 32.91 & 0.916 & 33.34 & 0.927 & 3.64 & 141.28 \\
MAXIM~\cite{tu2022maxim}           & 30.81 & 0.904 & 38.06 & 0.977 & 33.80 & 0.943 & 32.37 & 0.922 & 33.76 & 0.937 & 6.10 & 93.60 \\
Restormer~\cite{zamir2022restormer}       & 31.46 & 0.904 & 38.99 & 0.978 & 34.18 & 0.944 & 33.19 & 0.926 & 34.46 & 0.938 & 26.10 & 140.99 \\
IDT~\cite{IDT}             & 29.95 & 0.898 & 37.01 & 0.971 & 33.38 & 0.929 & 31.38 & 0.913 & 32.93 & 0.928 & 16.00 & 61.90 \\
IRNext~\cite{cui2023irnext}          & 31.64 & 0.902 & 38.24 & 0.972 & -     & -     & -     & -     & -     & -     & 13.21 & 114.00 \\
IR-SDE~\cite{IRSDE}          & 31.65 & 0.904 & 38.30 & 0.980 & 30.42 & 0.891 & -     & -     & -     & -     & 135.30 & 119.10 \\
DRSformer~\cite{DRSFormer}       & 31.66 & 0.901 &\underline{39.23} & 0.979 & 34.19 & 0.943 & 33.34 & 0.926 & 34.61 & 0.937 & 33.70 & 60.74 \\
MFDNet~\cite{MFDNetwang2023multi}          & 30.48 & 0.899 & 37.61 & 0.973 & 33.55 & 0.939 & 33.01 & 0.925 & 33.66 & 0.934 & 4.74 & 68.71 \\
MambaIR~\cite{guo2024mambair}         & 30.62 & 0.893 & 38.78 & 0.977 & 33.58 & 0.927 & 32.56 & 0.923 & 33.89 & 0.930 & 31.51 & 80.64 \\
VMambaIR~\cite{shi2024vmambair}        & 31.66 & 0.909 & 39.09 & 0.979 & 34.01 & 0.944 & 33.33 & 0.926 & 34.52 & 0.940 & -     & - \\
FreqMamba~\cite{zou2024freqmamba}       &\underline{31.74} &\underline{0.912} & 39.18 &\underline{0.981} &\underline{34.25} & \textbf{0.951} &\underline{33.36} &\underline{0.931} &\underline{34.63} &\underline{0.944} & 14.52 & 36.49 \\
\rowcolor{cyan!5} 
Ours            & \textbf{33.03} & \textbf{0.930} & \textbf{39.53} & \textbf{0.984} & \textbf{34.37} &\underline{0.944} & \textbf{34.93} & \textbf{0.938} & \textbf{35.47} & \textbf{0.949} & 2.80 & 25.16 \\
\bottomrule
\end{tabular}
\end{table*}

\subsection{Single Image Super Resolution}
\begin{figure*}[!t]
  \centering
  \includegraphics[width=1\linewidth]{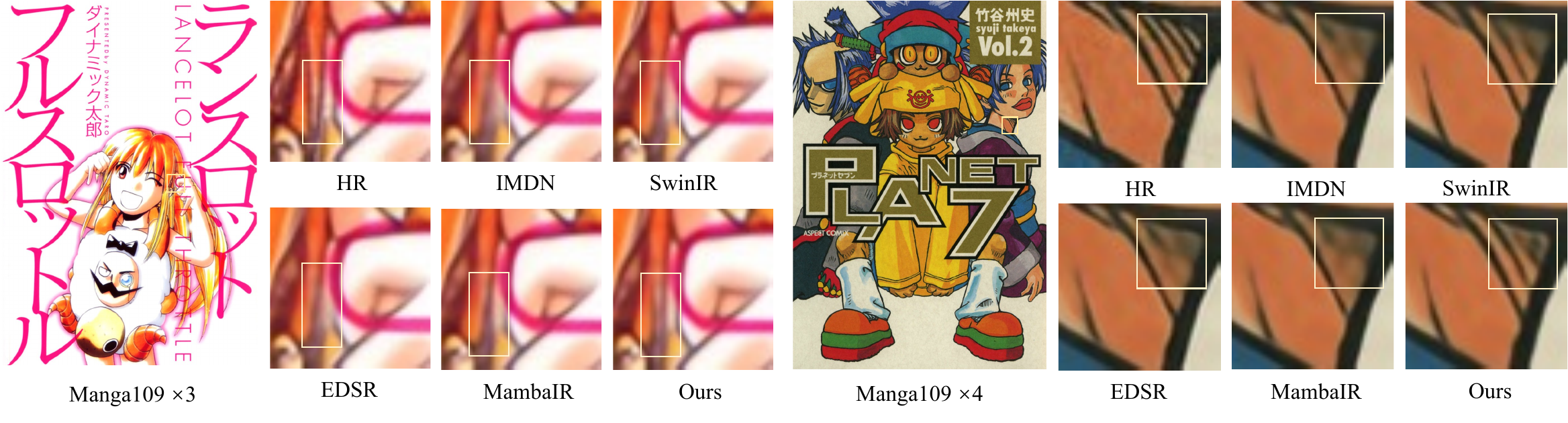}
  \caption{Visual comparison with general and specialized super-resolution methods under lightweight settings.}
  \label{fig:SR}
\end{figure*}

In the single-image super-resolution, we evaluate the performance of RestorMixer under lightweight settings with upscaling factors of $2\times$, $3\times$, and $4\times$. All models are trained on the DIV2K~\cite{DIV2K} dataset and evaluated on five widely-used benchmarks with diverse content ranging from simple patterns to complex textures. As shown in Tab.~\ref{tab:sr_comparison}, RestorMixer achieves competitive performance compared with both general and specialized SR methods. Although some SR expert methods, such as SRFormer-light~\cite{zhou2023srformer} achieve the best results on individual benchmarks, our method strikes a better balance between performance and efficiency. Visual comparisons in Fig.~\ref{fig:SR} further support this observation: our method effectively restores high-frequency textures and details, producing clean and natural results even in challenging regions.

\begin{table*}[!t]
\centering
\caption{Super-resolution performance comparison on benchmark datasets.\label{tab:sr_comparison}}

\renewcommand{\arraystretch}{1.2}
\setlength{\tabcolsep}{1.2pt}
\begin{tabular}{lcccccccccccccc}
\toprule
\multirow{2}{*}{\textbf{Methods}} & \multirow{2}{*}{\textbf{Scale}} & \multirow{2}{*}{\textbf{\#Param}} & \multirow{2}{*}{\textbf{MACs}} & \multicolumn{2}{c}{\textbf{Set5}} & \multicolumn{2}{c}{\textbf{Set14}} & \multicolumn{2}{c}{\textbf{BSDS100}} & \multicolumn{2}{c}{\textbf{Urban100}} & \multicolumn{2}{c}{\textbf{Manga109}} \\
 & & & & \textbf{PSNR} & \textbf{SSIM} & \textbf{PSNR} & \textbf{SSIM} & \textbf{PSNR} & \textbf{SSIM} & \textbf{PSNR} & \textbf{SSIM} & \textbf{PSNR} & \textbf{SSIM} \\
\midrule
CARN~\cite{CARN} & $\times2$ & 1,592K & 222.8G & 37.76 & 0.9590 & 33.52 & 0.9166 & 32.09 & 0.8978 & 31.92 & 0.9256 & 38.36 & 0.9765 \\
IMDN~\cite{IMDN} & $\times2$ & 694K & 158.8G & 38.00 & 0.9605 & 33.63 & 0.9177 & 32.19 & 0.8996 & 32.17 & 0.9283 & 38.88 & 0.9774 \\
LAPAR-A~\cite{li2020lapar} & $\times2$ & 548K & 171.0G & 38.01 & 0.9605 & 33.62 & 0.9183 & 32.19 & 0.8999 & 32.10 & 0.9283 & 38.67 & 0.9772 \\
LatticeNet~\cite{LatticeNet} & $\times2$ & 756K & 169.5G & 38.15 & 0.9610 & 33.78 & 0.9193 & 32.25 & 0.9005 & 32.43 & 0.9302 & - & - \\
SwinIR-light~\cite{liang2021swinir_swinir} & $\times2$ & 910K & 122.2G & 38.14 & 0.9611 & 33.86 & 0.9206 & 32.31 & 0.9012 & 32.76 & 0.9340 & 39.12 & 0.9783 \\
SRFormer-light~\cite{zhou2023srformer} & $\times2$ & 853K & 236G & \underline{38.23} & \underline{0.9613} & 33.94 & \underline{0.9209} & \textbf{32.36} & \textbf{0.9019} & \underline{32.91} & \underline{0.9353} & \underline{39.28} & \textbf{0.9785} \\
MambaIR~\cite{guo2024mambair} & $\times2$ & 905K & 167.1G & 38.13 & 0.9610 & \underline{33.95} & 0.9208 & 32.31 & 0.9013 & 32.85 & 0.9349 & 39.20 & \underline{0.9782} \\
\rowcolor{cyan!5} Ours & $\times2$ & 1066K & 93.7G & \textbf{38.22} & \textbf{0.9614} & \textbf{33.96} & \textbf{0.9217} & \underline{32.33} & \underline{0.9017} & \textbf{32.94} & \textbf{0.9358} & \textbf{39.33} & \underline{0.9782} \\
\midrule
CARN~\cite{CARN} & $\times3$ & 1,592K & 118.8G & 34.29 & 0.9255 & 30.29 & 0.8407 & 29.06 & 0.8034 & 28.06 & 0.8493 & 33.50 & 0.9440 \\
IMDN~\cite{IMDN} & $\times3$ & 703K & 71.5G & 34.36 & 0.9270 & 30.32 & 0.8417 & 29.09 & 0.8046 & 28.17 & 0.8519 & 33.61 & 0.9445 \\
LAPAR-A~\cite{li2020lapar} & $\times3$ & 544K & 114.0G & 34.36 & 0.9267 & 30.34 & 0.8421 & 29.11 & 0.8054 & 28.15 & 0.8523 & 33.51 & 0.9441 \\
LatticeNet~\cite{LatticeNet} & $\times3$ & 765K & 76.3G & 34.53 & 0.9281 & 30.39 & 0.8424 & 29.15 & 0.8059 & 28.33 & 0.8538 & - & - \\
SwinIR-light~\cite{liang2021swinir_swinir} & $\times3$ & 918K & 55.4G & 34.62 & 0.9289 & 30.54 & 0.8463 & 29.20 & 0.8082 & 28.66 & 0.8624 & 33.98 & 0.9478 \\
SRFormer-light~\cite{zhou2023srformer} & $\times3$ & 861K & 105G & \underline{34.67} & \underline{0.9296} & \underline{30.57} & \underline{0.8469} & \underline{29.26} & \textbf{0.8099} & \textbf{28.81} & \textbf{0.8655} & \underline{34.19} & \underline{0.9489}
 \\
MambaIR~\cite{guo2024mambair} & $\times3$ & 913K & 74.5G & 34.63 & 0.9288 & 30.54 & 0.8459 & 29.23 & 0.8084 & 28.70 & 0.8631 & 34.12 & 0.9479 \\
\rowcolor{cyan!5} Ours & $\times3$ & 1074K & 43.0G & \textbf{34.70} & \textbf{0.9298} & \textbf{30.65} & \textbf{0.8476} & \textbf{29.28} & \textbf{0.8099} & \textbf{28.81} & \underline{0.8654} & \textbf{34.37} & \textbf{0.9492} \\
\midrule
CARN~\cite{CARN} & $\times4$ & 1,592K & 90.9G & 32.13 & 0.8937 & 28.60 & 0.7806 & 27.58 & 0.7349 & 26.07 & 0.7837 & 30.47 & 0.9084 \\
IMDN~\cite{IMDN} & $\times4$ & 715K & 40.9G & 32.21 & 0.8948 & 28.58 & 0.7811 & 27.56 & 0.7353 & 26.04 & 0.7838 & 30.45 & 0.9075 \\
LAPAR-A~\cite{li2020lapar} & $\times4$ & 659K & 94.0G & 32.15 & 0.8944 & 28.61 & 0.7818 & 27.61 & 0.7366 & 26.14 & 0.7871 & 30.42 & 0.9074 \\
LatticeNet~\cite{LatticeNet} & $\times4$ & 777K & 43.6G & 32.30 & 0.8962 & 28.68 & 0.7830 & 27.62 & 0.7367 & 26.25 & 0.7873 & - & - \\
SwinIR-light~\cite{liang2021swinir_swinir} & $\times4$ & 930K & 31.8G & 32.44 & 0.8976 & 28.77 & 0.7858 & 27.69 & 0.7406 & 26.47 & 0.7980 & 30.92 & 0.9151 \\
SRFormer-light~\cite{zhou2023srformer} & $\times4$ & 873K & 62.8G & \underline{32.51} & \underline{0.8988} & \underline{28.82} & \underline{0.7872} & \textbf{27.73} & \textbf{0.7422} & \textbf{26.67} & \textbf{0.8032} & \textbf{31.17} & \textbf{0.9165} \\
MambaIR~\cite{guo2024mambair} & $\times4$ & 924K & 42.3G & 32.42 & 0.8977 & 28.74 & 0.7847 & 27.68 & 0.7400 & 26.52 & 0.7983 & 30.94 & 0.9135 \\
\rowcolor{cyan!5} Ours & $\times4$ & 1085K & 23.9G & \textbf{32.54} & \textbf{0.8992} & \textbf{28.85} & \textbf{0.7876} & \underline{27.72} & \textbf{0.7422} & \underline{26.60} & \underline{0.8013} & \textbf{31.17} & \underline{0.9163} \\
\bottomrule
\end{tabular}
\end{table*}

\subsection{Single Image De-snowing}
In the single image desnowing task, we evaluated the performance of RestorMixer on the CSD (2000)~\cite{csd} dataset. The experimental results are shown in Tab.~\ref{table:csd}, where RestorMixer achieves the best restoration results in terms of PSNR and SSIM, reaching 39.88dB and 0.99, significantly outperforming other methods. 

\begin{table}[!h]
\centering
\caption{Performance comparison of different methods on the CSD for snow removal.}
\label{table:csd}
\setlength{\tabcolsep}{8.6pt}
\renewcommand{\arraystretch}{1.2}
\begin{tabular}{lcccc}
\toprule
\textbf{Methods} & \textbf{PSNR} & \textbf{SSIM} & \textbf{Params(M)} & \textbf{FLOPs(G)} \\  
\midrule
DesnowNet~\cite{liu2018desnownet}  & 20.13 & 0.81 & 15.6 & 1.7K \\  
HDCW-Net~\cite{HDCW-Net}  & 29.06 & 0.91 & 6.99 & 9.78 \\  
SMGARN~\cite{SMGARN}  & 31.93 & 0.95 & 6.86 & 450.30 \\  
TransWeather~\cite{valanarasu2022transweather}  & 31.76 & 0.93 & 21.90 & 5.64 \\  
MPRNet~\cite{MPRNet}  & 33.98 & 0.97 & 3.64 & 141.28 \\  
DGUNet~\cite{DGUNet}  & 34.74 & 0.97 & 12.18 & 199.74 \\  
Uformer~\cite{wang2022uformer}  & 33.80 & 0.96 & 9.03 & 19.82 \\  
Restormer~\cite{zamir2022restormer}  & 35.43 & 0.97 & 26.10 & 140.99 \\  
NAFNet~\cite{nafnet}  & 35.13 & 0.97 & 22.40 & 12.12 \\   
SnowFormer~\cite{chen2022snowformer} &\underline{39.45} &\underline{0.98} & 8.38 & 19.44 \\  
MB-TaylorFormer~\cite{qiu2023mb} & 37.10 & 0.98 & 2.68 & 38.5 \\
\rowcolor{cyan!5}
Ours & \textbf{39.88} & \textbf{0.99} & 2.80 & 25.16 \\
\bottomrule
\end{tabular}
\end{table}

\subsection{Mixed Degradation Image Restoration}
\begin{figure}[t]
  \centering
  \includegraphics[width=.69\linewidth]{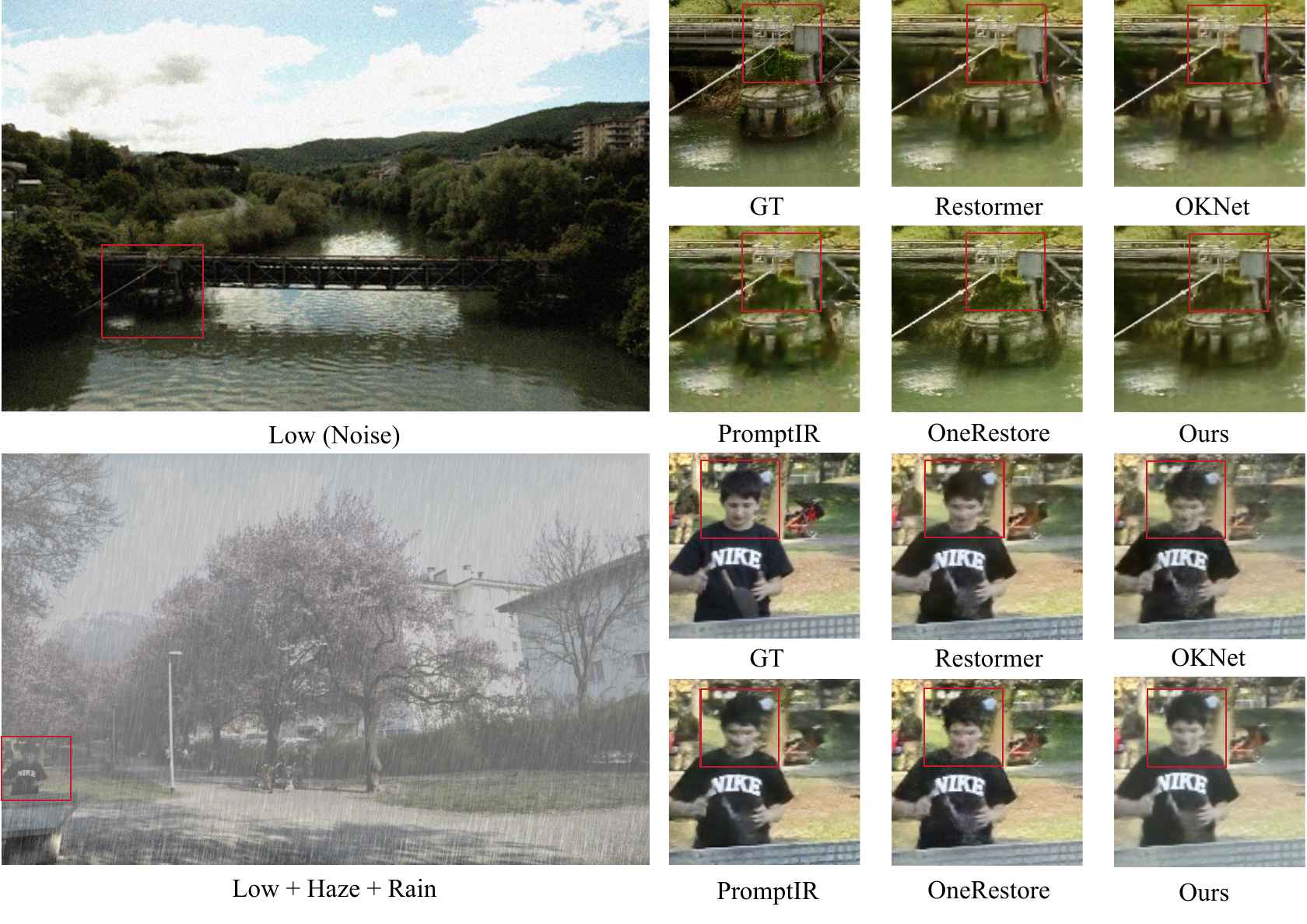}
  \caption{Visual comparison under low-light and composite degradations (Low+Haze+Rain). 
  }
  \label{fig:CDD}
\end{figure}

\begin{table}[!t]
\centering

\setlength{\tabcolsep}{7.1pt}
\renewcommand{\arraystretch}{1.2}
\caption{Performance comparison of different methods on mixed degradation image restoration.}
\label{table:CDD}
\begin{tabular}{clccc}
\toprule
\textbf{Types} & \textbf{Methods} & \textbf{PSNR} & \textbf{SSIM}& \textbf{Params(M)}\\ 
\midrule
\multirow{8}{*}{One-to-One} 
 & MIRNet~\cite{MIRNet} & 25.97 & 0.8474 &  31.79 \\  
 & MPRNet~\cite{MPRNet} & 25.47 & 0.8555 &  15.74 \\  
 & Restormer~\cite{zamir2022restormer} & 26.99 & 0.8646 & 26.99\\  
 & DGUNet~\cite{DGUNet} & 26.92 & 0.8559 & 17.33\\  
 & NAFNet~\cite{nafnet} & 24.13 & 0.7964 & 24.13\\  
 & SRUDC~\cite{srudc} & 27.64 & 0.8600 & 6.80\\  
 & OKNet~\cite{oknet} & 26.33 & 0.8605 & 4.72\\  
\midrule
\multirow{5}{*}{One-to-Many} 
 & AirNet~\cite{airnet} & 23.75 & 0.8140 & 8.93\\  
 & TransWeather~\cite{valanarasu2022transweather} & 23.13 & 0.7810 & 21.90\\  
 & WeatherDiff~\cite{weatherdiff} & 22.49 & 0.7985 & 82.96\\  
 & PromptIR~\cite{potlapalli2023promptir} & 25.90 & 0.8499 & 38.45\\  
 & WGWSNet~\cite{WGWSNet} & 26.96 & 0.8626 & 25.76\\  
\midrule
\multirow{1}{*}{One-to-Composite} 
 & OneRestore~\cite{guo2024onerestore} & \textbf{28.47} &\underline{0.8784} & 5.98\\ 
\midrule
\rowcolor{cyan!5} 
One-to-One & Ours &\underline{28.01} & \textbf{0.8807} & 2.80\\
\bottomrule
\end{tabular}
\end{table}
We evaluate RestorMixer on the CDD-11~\cite{guo2024onerestore} dataset for mixed degradation restoration. All models are trained on the full 11-subset training set using a unified strategy, without per-task tuning. Competing methods are categorized into one-to-one, one-to-many, and one-to-composite according to their adaptation mechanisms. As shown in Tab.~\ref{table:CDD}, RestorMixer achieves strong performance using a simple and general architecture without task-specific designs, scheduling strategies, or adaptive modules. It outperforms all one-to-one and one-to-many models, and performs on par with OneRestore~\cite{guo2024onerestore} — despite the latter being tailored for composite degradations. Notably, RestorMixer achieves the highest SSIM of 0.8807, indicating superior structural fidelity.
With fewer parameters and competitive results, RestorMixer offers a better trade-off between performance and efficiency. 
The visual results in Fig.~\ref{fig:CDD} further confirm its effectiveness in handling complex degradations.

\section{Ablation Study}

To evaluate the effectiveness of each design in RestorMixer, we perform ablation studies on Rain100L~\cite{100HL} and Rain100H~\cite{100HL} under identical training settings. Evaluation metrics include PSNR and SSIM for restoration quality, FLOPs (G) and Params (M) for complexity, and inference speed (ms) and memory usage for practical efficiency. The baseline model adopts a 3-level encoder-decoder architecture, with 4 hybrid blocks per layer and multi-scale attention windows of $8 \times 8$ and $16 \times 16$. 
Inference speed is measured as the average over 100 $256 \times 256$ patches, and memory refers to peak GPU usage during inference.

\subsection{Impact of Block Depth}

We investigate the effect of the number of stacked hybrid blocks in each encoder/decoder stage, comparing 2, 4 (baseline), and 6 blocks. Results in Table~\ref{tab:ablation-depth} show that increasing from 2 to 4 blocks improves both PSNR and SSIM, particularly on Rain100H with more complex rain streaks, highlighting the benefit of deeper architectures for multi-scale learning. However, increasing to 6 blocks provides little improvement while significantly increasing computational cost, memory usage, and latency. Therefore, 4 blocks strike the best balance between performance and efficiency.

\subsection{Effect of Multi-Scale Window Attention}


We evaluate the effectiveness of Multi-scale Window Self-attention (MWSA) by comparing the default multi-scale configuration ($8 \times 8 \to 16 \times 16$) with two fixed-size alternatives: all $8 \times 8$ and all $16 \times 16$. Results in Table~\ref{tab:ablation-windowsize} show significant performance drops with fixed window sizes, especially on Rain100H. This demonstrates that fixed receptive fields fail to adapt to spatial variations, limiting the capture of both local textures and global semantics. In contrast, the hierarchical window setting allows more flexible context aggregation, improving robustness to diverse rain patterns.

\subsection{Component-Level Structure Ablation}

We investigate the role of key components through four controlled experiments, each modifying or removing a specific part of the baseline model. Results are shown in Table~\ref{tab:ablation-components}.

The \textbf{``NoDSM''} variant removes downsampling in EMVM, keeping all operations at full resolution. While it retains detailed spatial information, it suffers from high computational cost and redundancy, demonstrating that long-range modeling is more efficient at lower resolutions. In the \textbf{``NoRDCNN''} variant, we replace the high-resolution RDCNN with the same M-T blocks. This introduces architectural uniformity but slightly degrades fine detail restoration, highlighting the importance of local convolutional priors in early decoding stages. The \textbf{``NoMWSA''} variant removes MWSA, replacing it with pure RDCNN in hybrid blocks. Performance, especially on Rain100H, drops significantly, showing that MWSA is crucial for capturing dynamic spatial structures. Unlike static CNN kernels, window attention adapts to the input, improving model flexibility. The \textbf{``FullTA''} variant replaces window self-attention with the transposed attention module from Restormer~\cite{zamir2022restormer}. Although it enables global modeling at channel dimension, it lacks spatial selectivity compared to MWSA and incurs higher computational cost, resulting in a poorer efficiency-performance trade-off.

\begin{table}[!t]
\centering

\caption{Ablation on number of blocks per stage.}
\setlength{\tabcolsep}{4.3pt}
\renewcommand{\arraystretch}{1.2}
\label{tab:ablation-depth}
\begin{tabular}{ccccccccc}
\toprule
\multirow{2}{*}{\textbf{\#Blocks}} & \multicolumn{2}{c}{Rain100L} & \multicolumn{2}{c}{Rain100H} & Params & FLOPs & Speed & Memory \\
& PSNR & SSIM & PSNR & SSIM & (M) & (G) & (ms) & (G) \\
\midrule
2&38.89&0.983&32.01&0.918&1.76&16.92&11.41&1.19\\
\rowcolor{cyan!5} 
4&39.53&0.984&\textbf{33.03}&\textbf{0.930}&2.80&25.16&20.55&2.22\\
6 &\textbf{39.58}&\textbf{0.985}&32.86&0.929&3.85&34.32&29.87&3.82\\
\bottomrule
\end{tabular}
\end{table}

\begin{table}[!t]
\centering

\caption{Ablation on window size setting in MWSA.}
\label{tab:ablation-windowsize}
\setlength{\tabcolsep}{3.9pt}
\renewcommand{\arraystretch}{1.2}
\begin{tabular}{ccccccccc}
\toprule
\textbf{\#Window} & \multicolumn{2}{c}{Rain100L} & \multicolumn{2}{c}{Rain100H} & Params & FLOPs & Speed & Memory \\
\textbf{Size} & PSNR & SSIM & PSNR & SSIM & (M) & (G) & (ms) & (G) \\
\midrule
\rowcolor{cyan!5}
\textbf{($8\to16$)}&\textbf{39.53}&\textbf{0.984}&\textbf{33.03}&\textbf{0.930}&2.80&25.16&20.55&2.22\\
$8\times8$ &39.29&0.984&32.14&0.920&2.80&25.05&20.53&2.05\\
$16\times16$ &39.08&0.984&32.05&0.919&2.80&25.26&20.63&2.35\\
\bottomrule
\end{tabular}
\end{table}

\begin{table}[!t]
\centering

\caption{Ablation on structural components of RestorMixer.}
\label{tab:ablation-components}
\renewcommand{\arraystretch}{1.2}
\setlength{\tabcolsep}{3.5pt}
\begin{tabular}{lcccccccc}
\toprule
\multirow{2}{*}{\textbf{\#Variant}} & \multicolumn{2}{c}{Rain100L} & \multicolumn{2}{c}{Rain100H} & Params & FLOPs & Speed & Memory \\
& PSNR & SSIM & PSNR & SSIM & (M) & (G) & (ms) & (G) \\
\midrule
\rowcolor{cyan!5}
Baseline&3\textbf{9.53}&\textbf{0.984}&\textbf{33.03}&\textbf{0.930}&2.80&25.16&20.55&2.22\\
NoDSM&39.32&0.984&32.62&0.925&2.76&27.56&25.66&2.78\\
NoRDCNN&39.44&0.984&32.66&0.927&2.83&25.89&31.51&5.28 \\
NoMWSA&38.91&0.983&30.28&0.902&2.86&25.41&18.17&1.81\\
FullTA&38.82&0.983&32.30&0.921&2.80&24.98&28.52&2.04\\
\bottomrule
\end{tabular}
\end{table}

\section{Conclusion}

In this paper, we address the limitation in performance improvement of current models for various IR tasks, which are typically designed based on a single underlying architecture. We propose a novel IR model named RestorMixer, which is meticulously constructed by integrating three heterogeneous underlying architectures. Specifically, RestorMixer capitalizes on the individual strengths of these three structures: it employs RDCNN block for its efficient parallel computation and ability to extract shallow features, to process the highest-resolution feature maps in the model's encoding/decoding stages; it utilizes EMVM block to rapidly model global feature dependencies, and optimizes the computational redundancy and memory attenuation issues present in the classical multi-directional SSM scanning mechanism; and it further refines local feature extraction and representation using multi-scale window self-attention. Extensive experimental validation demonstrates that the proposed RestorMixer achieves leading performance on multiple image degradation restoration benchmark datasets, including single degradation and composite degradation IR, while also exhibiting highly efficient inference speed.

\bibliographystyle{unsrt}
\bibliography{sample-base}

\end{document}